\documentclass{article}
\usepackage{spconf,amsmath,amssymb,graphicx,booktabs,hyperref}
\usepackage{xcolor}
\usepackage{array}
\newcommand{\benchname}{SLP-Bench}

\newcommand{\icon}[1]{\raisebox{-0.35ex}{\includegraphics[height=2.3ex]{fig/icons/#1.pdf}}}
\newcommand{\unsection}[1]{\par\vspace{1.2ex}{\centering\bfseries\MakeUppercase{#1}\par}\vspace{0.8ex}}
\newcommand{\flag}[1]{\raisebox{0.1ex}{\includegraphics[height=1.3ex]{fig/icons/flag_#1.pdf}}}

\makeatletter
\let\spc@makecaption\@makecaption
\long\def\@makecaption#1#2{\small\spc@makecaption{#1}{#2}}
\makeatother

\title{SLP-Bench: Evaluating and Improving Acoustic Claims of Reasoning Audio Language Models on Clinical Speech}

\name{C\^{o}me Hosxe \qquad Rachid Riad \qquad Quentin Spinat}
\address{Callyope, Paris, France}

\begin{document}
\ninept
\maketitle

\begin{abstract}
Audio language models (ALMs) can answer clinical questions from speech, outperforming baselines built on handcrafted acoustic features. Reasoning ALMs generate a chain of thought that refers to properties of the speaker's voice before predicting. Each statement the model makes about the voice needs to be checked against clinicians' perceptual ratings of the same recording. On clinical speech, existing evaluations score only the final answer. We introduce Speech Language Pathologist Bench (\benchname{}), a benchmark grading each acoustic claim in the chain of thought against clinicians' perceptual ratings of the same recording. It covers five clinician-rated datasets and 25 perceptual dimensions, and assesses claims for factuality, stability under resampling, and consistency across reworded and different questions. Across ten open-weight reasoning ALMs evaluated zero-shot, claim balanced accuracy ranges from 0.52 to 0.58, claims change under resampling and rewording, and the most consistent models are not the most factual. Fine-tuning a 7B ALM with GRPO on a reward for claims that agree with clinicians raises its claim balanced accuracy from 0.54 to 0.60 and its mean $\kappa$ from 0.39 to 0.61, while preserving answer accuracy on per-dimension questions. \benchname{} is available upon request, subject to approval by the providers of each dataset.
\end{abstract}


\begin{figure*}[t]
\centering
\includegraphics[width=\textwidth]{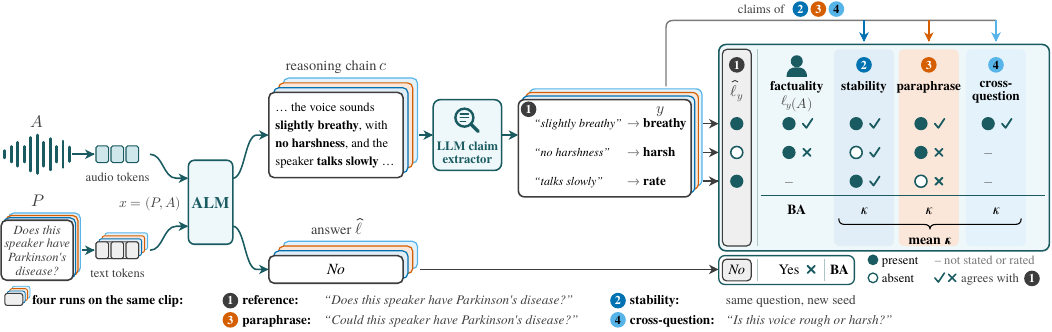}
\caption{\benchname{}: claims extracted from the chain of thought are graded against clinician ratings and compared across runs}\label{fig:pipeline}
\end{figure*}

\section{Introduction}
\label{sec:intro}
Audio language models (ALMs) have the potential to become the state-of-the-art in clinical applications of speech processing, such as the diagnosis of Parkinson's disease or amyotrophic lateral sclerosis~\cite{amiri2026}. Reasoning ALMs such as Qwen3-Omni~\cite{qwen3omni}, Step-Audio-R1~\cite{stepaudior1}, Nemotron-3 Nano Omni~\cite{nemotronomni}, MiniCPM-o 4.5~\cite{minicpmo45}, MOSS-Audio~\cite{mossaudio}, Audio Flamingo 3 and Next~\cite{af3,afnext}, Gemma 4~\cite{gemma4} and Audio-Reasoner~\cite{audioreasoner} generate a chain of thought before answering. Asked whether a speaker has Parkinson's disease, such a model may state that the voice is breathy, monotonous and of reduced loudness, and only then give its prediction. These are perceptual dimensions that clinicians rate on standardized scales~\cite{darley1969,capev,grbas}. If they are stated correctly, the chain gives clinicians an explanation they can check against their own perceptual judgment.

Current audio reasoning benchmarks score only the final answer~\cite{mmau,mmar,mmsu,patel2025}. Other work evaluates the chain itself, with an LLM judge applying rubrics derived from reference chains~\cite{mmarrubrics}, with step-level labels from LLM and human annotators~\cite{audioprocessbench}, or with an ALM that verifies each statement~\cite{emosura}. In these studies, the reference is a human or model judgment on general audio or emotional speech, not a clinician rating. Whether ALMs rely on acoustic cues at all is itself uncertain: for emotion, they depend mainly on lexical content~\cite{chen2026listen}. For clinical speech, Chen et al.~\cite{chen2026} used an ALM to predict CAPE-V and GRBAS scores as input to a downstream classifier, without free-form reasoning.

A chain of thought does not necessarily reflect how the model reaches its answer~\cite{turpin2023,lanham2023,jacovi2020}, but its statements about the voice can still be checked: against the clinician's rating of the same recording (factuality), and across new chains and questions on that recording (stability and consistency). To our knowledge, measuring these properties for reasoning ALMs on clinical speech remains largely underexplored.

We make three contributions. (i)~Speech Language Pathologist Bench (\benchname{}), a benchmark that grades the acoustic claims of a reasoning chain against clinician ratings of the same recording and scores them for factuality, stability and consistency (Section~\ref{sec:bench}). We apply it to five clinician-rated datasets, with 794 test clips from 260 patients and 3,452 prompts at three levels of specificity, on 25 perceptual dimensions grouped by speech production system (Section~\ref{sec:setup}). (ii)~A zero-shot evaluation of ten open-weight reasoning ALMs on the factuality, stability and consistency of their claims, and on the clinical balanced accuracy of their answers (Section~\ref{sec:results}). (iii)~GRPO training of a 7B ALM with a reward on claim factuality, compared with rewards on the answer alone and on both (Section~\ref{sec:training}); the claim reward makes the model both more factual and more consistent than every zero-shot ALM.

\section{\benchname{}}
\label{sec:bench}

We introduce \benchname{}, a benchmark that tests whether the acoustic claims a reasoning ALM makes about a recording are correct, stable across resampled outputs, and consistent across related prompts. It proceeds in three steps (Fig.~\ref{fig:pipeline}). 
First, the model receives a prompt and a recording and generates a chain of thought followed by an answer (Section~\ref{ssec:notation}). 
Second, an LLM-based extractor maps the chain of thought onto claims about perceptual dimensions, and grades each claim against the clinician's rating of that dimension, used as gold standard (Section~\ref{sec:grading}). 
Third, graded claims are scored for factuality, stability and consistency, and answers by balanced accuracy (BA) on the clinical endpoint (Section~\ref{ssec:metrics}). 
We define these steps in general terms here; Section~\ref{sec:setup} describes the instance we evaluate.

\subsection{Inputs, dimensions and targets}
\label{ssec:notation}
An input $x=(P,A)$ consists of a prompt $P$ and a recording $A$. 
Given $x$, the model generates a chain of thought $c$ followed by an answer $\hat{\ell}\in\{0,1\}$. 
Let $\mathcal{Y}$ denote the set of perceptual dimensions, i.e., properties of speech that clinicians judge by ear. 
Let $\mathcal{R}(A)\subseteq\mathcal{Y}$ denote the subset of dimensions rated by a clinician on recording $A$, which depends on the rating scale of its dataset. 
For each $y\in\mathcal{R}(A)$, the label $\ell_y(A)\in\{0,1\}$ equals~$1$ if the clinician judged dimension $y$ present and~$0$ otherwise. 
Similarly, let $\mathcal{Z}$ denote the set of clinical targets (e.g., diagnosis), with speaker-level labels $\ell_z(A)\in\{0,1\}$ for $z\in\mathcal{Z}$.

\subsection{Claims and grading}
\label{sec:grading}

Our LLM claim extractor reads only $c$, without access to the recording, the prompt or the labels, and returns the claims it makes about the voice. A claim $(y,\hat{\ell}_y)$ asserts that dimension $y\in\mathcal{Y}$ is present ($\hat{\ell}_y=1$) or absent ($\hat{\ell}_y=0$), with at most one claim per dimension. A claim is graded if its dimension was rated, i.e., $y\in\mathcal{R}(A)$, and is correct if it agrees with $\ell_y(A)$. We denote by $G(x)$ the set of graded claims for input $x$. Claims are extracted for any dimension in $\mathcal{Y}$, regardless of what the prompt asks.

\subsection{Metrics}
\label{ssec:metrics}

Graded claims are scored for factuality, stability and consistency, and answers for clinical balanced accuracy (BA).

\textbf{Factuality} is the BA of graded claims against the labels. BA is the standard metric in computational paralinguistics for imbalanced labels~\cite{schuller2013}, and equals 0.5 for a model that ignores the audio. We compute one BA per dataset and dimension, and average them weighted by their number of graded claims.

\textbf{Stability and consistency} are measured by Cohen's kappa between the values that two chains of thought on the same recording assign to the dimensions they both mention, pooled over recordings. Stability compares two chains for the same input sampled with different seeds, paraphrase consistency compares a prompt with its paraphrase, and cross-question consistency compares two prompts with different targets. We denote their average by $\kappa$.

\textbf{Clinical BA} is the balanced accuracy of answers against the label of the prompt's target, $\ell_y(A)$ or $\ell_z(A)$. It shows whether changes in the claims are reflected in the answers.

\section{Experimental setup}
\label{sec:setup}

\begin{table}[t]
\caption{The five datasets, test split. h: hours. Dur.: mean clip duration (s). $\mathcal{Y}$: rating scale and presence threshold. $\mathcal{Z}$: PD Parkinson's disease, path.\ pathological voice, UVFP unilateral vocal fold paralysis.}
\label{tab:xyz}
\vspace{2pt}
\centering\small\setlength{\tabcolsep}{2.2pt}
\begin{tabular}{@{}lrrr>{\raggedright\arraybackslash}p{2.15cm}l@{}}
\toprule
 & \multicolumn{3}{c}{$X$} & $\mathcal{Y}$ & $\mathcal{Z}$ \\
\cmidrule(lr){2-4}\cmidrule(lr){5-5}\cmidrule(l){6-6}
Dataset & Clips & h & Dur. & Scale, threshold & Target \\
\midrule
\flag{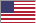} SAP~\cite{sap}             & 246 & 0.99 & 14.5 & Mayo $\geq 3$ & PD \\
\flag{us} HeyJay~\cite{heyjay}       & 95  & 0.15 & 5.6  & Mayo $\geq$ \emph{moderate} & PD \\
\flag{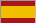} NeuroVoz~\cite{neurovoz}   & 55  & 0.20 & 12.9 & GRBAS $\geq 1$, note $\neq$ normal & PD \\
\flag{us} PVQD~\cite{pvqd}           & 50  & 0.31 & 22.0 & CAPE-V $\geq 15$, GRBAS $\geq 1$ & path. \\
\flag{us} Bridge2AI~\cite{bridge2ai} & 348 & 1.75 & 18.1 & CAPE-V $\geq 15$ & UVFP \\
\midrule
Total & 794 & 3.39 & 15.4 & & \\
\bottomrule
\end{tabular}
\end{table}

\subsection{Datasets}
\label{ssec:datasets}
We use five datasets that provide both clinician ratings of perceptual voice dimensions and the clinical status of each speaker (Table~\ref{tab:xyz}). SAP, HeyJay and NeuroVoz include speakers with motor speech disorders such as Parkinson's disease, amyotrophic lateral sclerosis, cerebral palsy or ataxia, while PVQD and Bridge2AI include speakers with voice disorders. All are in English except NeuroVoz (Spanish). SAP and HeyJay contain sentences of more than five words (up to five per patient); NeuroVoz a sustained vowel, a monologue and a diadochokinetic task; Bridge2AI a sustained vowel, a read passage and a maximum phonation task; and PVQD one recording per patient, with sustained vowels followed by the CAPE-V sentences. The test split comprises 260 patients. The remaining patient-disjoint training splits are not part of the benchmark and are used only to train our model (Section~\ref{sec:training}): 256 patients and 6,668 clips from SAP, 84 and 2,198 from NeuroVoz, 216 and 216 from PVQD, and 103 and 495 from Bridge2AI; HeyJay is test-only. Supervised fine-tuning additionally uses other speech datasets.

\subsection{Dimensions and labels}
\label{ssec:dimensions}

\begin{table}[t]
\caption{The 25 dimensions of $\mathcal{Y}$ by speech production system, the datasets that rate them (S SAP, H HeyJay, N NeuroVoz, P PVQD, B Bridge2AI) and \%, the share of rated test clips on which the dimension is present.}
\label{tab:dims}
\vspace{2pt}
\centering\small\setlength{\tabcolsep}{2.2pt}
\begin{tabular}{@{}llr@{\hspace{5pt}}llr@{}}
\toprule
Dimension & Data & \% & Dimension & Data & \% \\
\midrule
\multicolumn{3}{@{}l}{\icon{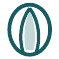}\,\emph{Phonation}} & \multicolumn{3}{l}{\icon{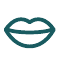}\,\emph{Articulation}} \\
harsh & SHNPB & 56 & imprecise consonants & SHN & 27 \\
breathy & SHNPB & 26 & distorted vowels & SH & 38 \\
strained-strangled & SHNPB & 43 & prolonged phonemes & SH & 1 \\
hoarse-wet & H & 0 & irregular breakdowns & SH & 5 \\
hard glottal attack & N & 7 & hypernasality & SHN & 14 \\
pitch level & SNPB & 33 & nasal emission & S & 0 \\
pitch breaks & SH & 3 & rate & SN & 15 \\
monopitch & SH & 54 & variable rate & SH & 8 \\
\multicolumn{3}{@{}l}{\icon{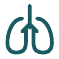}\,\emph{Respiration}} & short rushes & S & 9 \\
loudness level & SNPB & 50 & prolonged intervals & S & 32 \\
monoloudness & SH & 54 & inappropriate silences & SH & 4 \\
loudness decay & H & 5 & & & \\
excess loudness var. & S & 0 & & & \\
audible inspiration & SH & 2 & & & \\
short phrases & SH & 5 & & & \\
\bottomrule
\end{tabular}
\end{table}

The Mayo dysarthria items~\cite{darley1969}, CAPE-V~\cite{capev} and GRBAS~\cite{grbas} often name the same feature differently, so we define $\mathcal{Y}$ as a single list of 25 dimensions named after Darley et al.~\cite{darley1969} (Table~\ref{tab:dims}). We exclude three Mayo items that describe speech globally: overall severity, intelligibility and naturalness.
\label{ssec:mapping}
A mapping $\varphi$ assigns each rating item to a dimension of $\mathcal{Y}$; for instance, CAPE-V roughness, GRBAS roughness and the Mayo item \emph{harsh voice} all map to \emph{harsh}. A dimension is present on a recording ($\ell_y(A)=1$) when at least one item mapped to it reaches the threshold of its scale (Table~\ref{tab:xyz}), and $\mathcal{R}(A)$ holds the dimensions with at least one rated item. Bridge2AI and HeyJay provide session-level ratings, which apply to every clip of the session.
\vspace{-0.5em}

\subsection{Prompts}
\label{ssec:questions}
Clips receive three types of prompt. Simple questions, one per speech subsystem, ask \emph{Is this voice rough or harsh?}, \emph{Is the loudness of this voice abnormal (weak, monotone or decaying)?} and \emph{Is this speaker's articulation imprecise?}, only on clips rated on that dimension, in datasets where both labels occur. The clinical endpoint question asks about the dataset's target in $\mathcal{Z}$. The description request, \emph{Describe the voice and speech of this speaker.}, has no target. Yes/no questions end with \emph{Answer Yes or No.}; the answer is the first \emph{yes} or \emph{no} after the reasoning block. The test split holds 3,452 prompts: 1,864 simple, 794 endpoint and 794 description.

\vspace{-0.5em}

\subsection{LLM claim extractor}
\label{ssec:extractor}
\label{ssec:resextractor}
The LLM claim extractor is Qwen3-30B-A3B-Instruct-2507~\cite{qwen3} at temperature 0. A negation is a claim of absence, and a hedged statement or a plain description a claim of presence (\emph{seems strained}, \emph{a deep voice}). Statements on speech content, diagnosis or emotion are not acoustic claims. On 98 synthetic sentences written from a known claim or from no claim, including negations, hedged and hypothetical statements, and the question of the prompt repeated inside the chain, the extractor returned the expected output for 86 (accuracy 0.88). On 63 test chains with LLM-annotated claims, it reached a precision of 0.78, a recall of 0.61, and an F1 of 0.69.

\subsection{Evaluation protocol}
\label{ssec:protocol}
A clip is one speech task of one patient, cut at 30\,s and resampled to 16\,kHz. BA is computed only for dataset--dimension cells with at least 10 graded claims and both label. Stability uses 500 test questions resampled with a second seed, and paraphrase consistency 380 reworded yes/no questions. We report 95\% patient-level bootstrap intervals over 500 resamples, shared by all models. All models are served with vLLM at temperature 0.6, top-p 0.95, at most 4,096 new tokens and a fixed seed, one setting for all, sampled rather than greedy because stability compares two seeds.

\section{Models}
\label{sec:models}

\subsection{Baselines}
\label{ssec:models}
We evaluate ten open-weight reasoning ALMs in a zero-shot setting: Qwen3-Omni Thinking~\cite{qwen3omni}, Step-Audio-R1 and R1.1~\cite{stepaudior1}, Nemotron-3 Nano Omni~\cite{nemotronomni}, MiniCPM-o 4.5~\cite{minicpmo45}, MOSS-Audio-8B-Thinking~\cite{mossaudio}, Audio Flamingo Next Think~\cite{afnext}, Audio Flamingo 3 Think~\cite{af3}, Gemma-4-E4B~\cite{gemma4}, and Audio-Reasoner~\cite{audioreasoner}. All models receive the same prompts and clips in one user turn, and use the reasoning mode documented by their authors. As a reference, a chance model states each claim and each answer at random. Its expected BA and clinical BA are 0.5. Kappa is corrected for chance agreement by construction, so we give no chance value for stability and consistency.

\subsection{Our model}
\label{sec:training}
We test whether rewarding claim factuality makes a model's chains of thought more factual. We start from Qwen2.5-Omni 7B~\cite{qwen25omni}. We first fine-tune it with LoRA~\cite{lora} on chains generated by Qwen3-Omni Thinking~\cite{qwen3omni} on clinical and non-clinical speech, including the training splits of SAP, NeuroVoz and PVQD, keeping only chains whose answer and verified claims are correct. No test patient appears in these chains. We call this model SFT.

From SFT, we train three models with GRPO~\cite{grpo} that differ only in their reward:
\begin{equation}
r = w_\mathrm{v}\,r_\mathrm{verdict} + w_\mathrm{c}\,r_\mathrm{claim} + 0.2\,r_\mathrm{format},
\end{equation}
where $r_\mathrm{verdict}=1$ if the yes/no answer matches the label, and $r_\mathrm{format}=1$ if the output has one reasoning block followed by an answer. The claim term applies the grader of Section~\ref{sec:grading} to training clips:
\begin{equation}
r_\mathrm{claim} = \frac{1}{|\mathcal{R}(A)|}\sum_{y \in G(x)} \frac{h_y - p_y}{1 - p_y},
\end{equation}
where $h_y=1$ if the claim on $y$ is correct and 0 otherwise, and $p_y$ is the probability that its stated value is correct by chance, given by the training prevalence of $y$ in its dataset. A claim independent of the audio thus scores 0 in expectation, and a chain without graded claims receives $r_\mathrm{claim}=0$. The \emph{answer} model rewards only the answer ($w_\mathrm{v}=1, w_\mathrm{c}=0$), the \emph{claims} model only the claims ($w_\mathrm{v}=0, w_\mathrm{c}=3$), and the \emph{answer + claims} model both ($w_\mathrm{v}=1, w_\mathrm{c}=3$); the answer model isolates the effect of GRPO without a claim reward.

GRPO uses the rated clips of the SAP, NeuroVoz, PVQD and Bridge2AI training splits, each rating turned into a yes/no question. For each question, we sample 8 chains of at most 640 tokens and train rank-16 LoRA adapters with a learning rate of $2\cdot10^{-6}$. The three models are compared after 1,553 steps. As the claims model was still improving, we continued its training within our compute budget and also report its last checkpoint, at step 2,700.
\vspace{-0.5em}

\section{Results}
\label{sec:results}
\begin{figure}[t]
\centering
\includegraphics[width=\columnwidth]{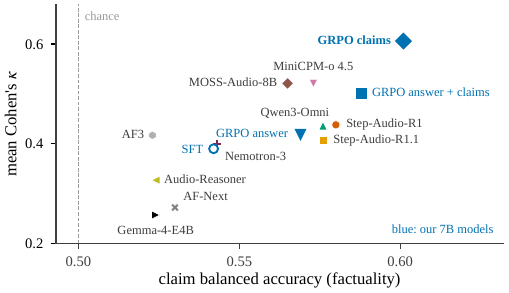}
\caption{Claim balanced accuracy (factuality) of ALM  compared to ground truth annotations against mean $\kappa$, the mean of stability, paraphrase and cross-question consistency, all prompts, test split.}
\label{fig:pareto}
\end{figure}

On the test split, the extractor graded 60,191 claims from the chains of 14 models on test sets of the five corpora: ten zero-shot reasoning ALMs, SFT and our three GRPO models. Zero-shot ALMs make claims about the voice that are barely above chance and change when the chain is resampled or the question reworded (Fig.~\ref{fig:pareto}, all $\kappa \leq 0.61$). Only the claims model is both more factual and more consistent than every zero-shot ALM (Fig.~\ref{fig:pareto}).

\noindent\textbf{Factuality vs.\ consistency.} Among zero-shot models, the two do not go together. The most factual, the two Step-Audio models and Qwen3-Omni Thinking, agree only moderately with their own claims across runs; the most consistent, MiniCPM-o 4.5 and MOSS-Audio-8B-Thinking, are less often correct; and AF-Next Think, Gemma-4-E4B and Audio-Reasoner are near chance in factuality and the least consistent. Rewarding the answer barely moves SFT, whereas rewarding the claims moves it beyond all zero-shot models on both axes, to a claim BA of 0.60 and a $\kappa$ of 0.61.

\noindent\textbf{Answer reward.} Rewarding the answer alone does not even improve the answers. The answer model learns to say \emph{no} to almost every clinical endpoint question (0.1\% \emph{yes}), which fixes its BA on these questions at 0.5, and its clinical BA on simple questions drops from 0.61 to 0.55. The claim reward avoids this shortcut: the claims model keeps the clinical BA of SFT on simple questions, suggesting that rewarding what the chain says about the voice also keeps the answers grounded.

\begin{figure}[t]
\centering
\includegraphics[width=\columnwidth]{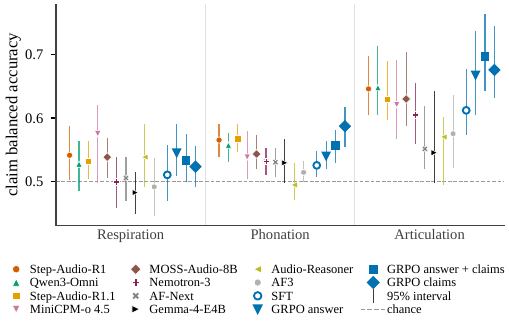}
\caption{Decomposition of claim factuality by step of the speech production system, all prompts, test split, with 95\% patient-bootstrap intervals.}
\label{fig:systems}
\end{figure}

\noindent\textbf{Speech production sub-analysis.} Factuality depends more on the subsystem than on the model (Fig.~\ref{fig:systems}). All models are most factual on articulation, where the two models trained with the claim reward lead. On phonation, zero-shot models stay slightly above chance and the claims model exceeds all of them. Respiratory dimensions, such as loudness and phrase length, remain near chance for every ALM, suggesting that current audio encoders capture them poorly.

\begin{figure}[t]
\centering
\includegraphics[width=\columnwidth]{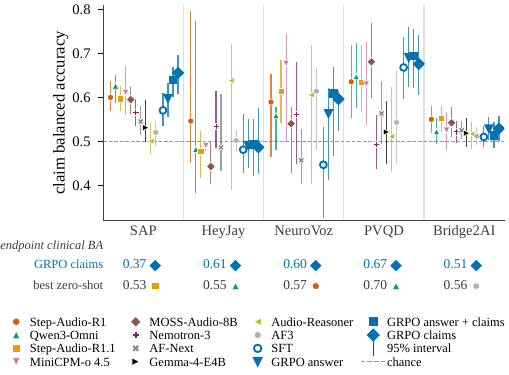}
\caption{Top: claim BA by dataset (all test prompts, 95\% patient-bootstrap intervals). Bottom: endpoint answer BA of the claims model and the best zero-shot model (marker).}
\label{fig:datasets}
\end{figure}

\noindent\textbf{Prompt type and datasets.} Claims are less factual when the prompt asks about a diagnosis: eight of the nine models with claims under both prompt types are less factual on endpoint questions than on simple questions (not shown). The claims model keeps its factuality on endpoint questions, where it is the most factual model, and it is the most factual on SAP, the largest dysarthria dataset (Fig.~\ref{fig:datasets}, top). Its endpoint answers also exceed those of the best zero-shot model on HeyJay and NeuroVoz, while Bridge2AI remains near chance for every model (Fig.~\ref{fig:datasets}, bottom).

\vspace{-0.5em}

\section{Conclusion}
\label{sec:conclusion}
We introduced \benchname{}, a benchmark that grades the acoustic claims in the chains of thought of reasoning ALMs against clinician ratings of the same recordings. Across five clinician-rated datasets, the claims of ten open reasoning ALMs reach a BA of only 0.52 to 0.58 and change under resampling and rewording, and the most consistent models are not the most factual: a model agreeing with itself is not evidence that it is right. These chains therefore cannot yet serve as explanations that clinicians check against their own perception. Rewarding verifiable claims narrows this gap. A GRPO reward on the claims raises the claim BA of a 7B model from 0.54 to 0.60 and its $\bar{\kappa}$ from 0.39 to 0.61 without lowering its clinical BA on simple questions, whereas a reward on the answer alone collapses to answering \emph{no}. Respiratory dimensions, however, remain near chance even with the claim reward, suggesting a limitation of the audio representations rather than of the reasoning. 
Future work includes a clinician-annotated gold set, equal training budgets across rewards, tuning the relative weights of rewards, and a dedicated study of whether more factual claims improve clinical endpoint prediction.

\clearpage
\unsection{Compliance with Ethical Standards}
This retrospective study uses human subject data from the five datasets of Section~\ref{ssec:datasets}. Ethical approval was not required, as confirmed by their licenses or data use agreements.

\unsection{Acknowledgments}
All authors are employees of Callyope, which funded this work and develops AI tools for psychiatry. This work was granted access to the HPC/AI resources of IDRIS under the allocation 20XX-AD010315777 made by GENCI. We acknowledge EuroHPC JU for awarding the project ID EHPC-DEV-2026D02-260 access to Leonardo hosted by CINECA, Italy.

\bibliographystyle{IEEEbib}
\bibliography{refs}

@article{darley1969,
  author  = {Darley, Frederic L. and others},
  title   = {Differential diagnostic patterns of dysarthria},
  journal = {J. Speech Hear. Res.},
  volume  = {12},
  number  = {2},
  pages   = {246--269},
  year    = {1969}
}

@article{capev,
  author  = {Kempster, Gail B. and others},
  title   = {Consensus auditory-perceptual evaluation of voice: development of a standardized clinical protocol},
  journal = {American Journal of Speech-Language Pathology},
  volume  = {18},
  number  = {2},
  pages   = {124--132},
  year    = {2009}
}

@book{grbas,
  author    = {Hirano, Minoru},
  title     = {Clinical Examination of Voice},
  publisher = {Springer},
  year      = {1981}
}

@article{sap,
  author  = {Hasegawa-Johnson, Mark and others},
  title   = {Community-supported shared infrastructure in support of speech accessibility},
  journal = {Journal of Speech, Language, and Hearing Research},
  volume  = {67},
  number  = {11},
  pages   = {4162--4175},
  year    = {2024},
  doi     = {10.1044/2024_JSLHR-24-00122}
}

@article{neurovoz,
  author  = {Mendes-Laureano, Jana{\'i}na and others},
  title   = {{NeuroVoz}: a {Castillian Spanish} corpus of parkinsonian speech},
  journal = {Scientific Data},
  volume  = {11},
  pages   = {1367},
  year    = {2024},
  doi     = {10.1038/s41597-024-04186-z}
}

@article{pvqd,
  author  = {Walden, Patrick R.},
  title   = {Perceptual Voice Qualities Database ({PVQD}): database characteristics},
  journal = {Journal of Voice},
  volume  = {36},
  number  = {6},
  pages   = {875.e15--875.e23},
  year    = {2022},
  doi     = {10.1016/j.jvoice.2020.10.001}
}

@misc{bridge2ai,
  author       = {Johnson, Alistair and others},
  title        = {{Bridge2AI-Voice}: an ethically-sourced, diverse voice dataset linked to health information},
  note         = {Version 1.1},
  howpublished = {PhysioNet},
  year         = {2025},
  doi          = {10.13026/249v-w155}
}

@article{qwen3,
  author  = {{Qwen Team}},
  title   = {Qwen3 technical report},
  journal = {arXiv preprint arXiv:2505.09388},
  year    = {2025}
}

@article{qwen25omni,
  author  = {Xu, Jin and others},
  title   = {{Qwen2.5-Omni} technical report},
  journal = {arXiv preprint arXiv:2503.20215},
  year    = {2025}
}

@article{grpo,
  author  = {Shao, Zhihong and others},
  title   = {{DeepSeekMath}: pushing the limits of mathematical reasoning in open language models},
  journal = {arXiv preprint arXiv:2402.03300},
  year    = {2024}
}

@inproceedings{lora,
  author    = {Hu, Edward J. and others},
  title     = {{LoRA}: low-rank adaptation of large language models},
  booktitle = {Proc. ICLR},
  year      = {2022}
}

@inproceedings{audioreasoner,
  author    = {Xie, Zhifei and others},
  title     = {{Audio-Reasoner}: improving reasoning capability in large audio language models},
  booktitle = {Proc. EMNLP},
  pages     = {23829--23851},
  year      = {2025},
  doi       = {10.18653/v1/2025.emnlp-main.1216}
}

@inproceedings{af3,
  author    = {Goel, Arushi and Ghosh, Sreyan and others},
  title     = {{Audio Flamingo 3}: advancing audio intelligence with fully open large audio language models},
  booktitle = {Proc. NeurIPS},
  year      = {2025}
}

@article{afnext,
  author  = {Ghosh, Sreyan and others},
  title   = {{Audio Flamingo Next}: next-generation open audio-language models for speech, sound, and music},
  journal = {arXiv preprint arXiv:2604.10905},
  year    = {2026}
}

@article{qwen3omni,
  author  = {{Qwen Team}},
  title   = {{Qwen3-Omni} technical report},
  journal = {arXiv preprint arXiv:2509.17765},
  year    = {2025}
}

@article{stepaudior1,
  author  = {Tian, Fei and others},
  title   = {{Step-Audio-R1} technical report},
  journal = {arXiv preprint arXiv:2511.15848},
  year    = {2025}
}

@inproceedings{mmau,
  author    = {Sakshi, S. and others},
  title     = {{MMAU}: a massive multi-task audio understanding and reasoning benchmark},
  booktitle = {Proc. ICLR},
  year      = {2025}
}

@inproceedings{mmar,
  author    = {Ma, Ziyang and others},
  title     = {{MMAR}: a challenging benchmark for deep reasoning in speech, audio, music, and their mix},
  booktitle = {Proc. NeurIPS Datasets and Benchmarks Track},
  year      = {2025}
}

@inproceedings{jacovi2020,
  author    = {Jacovi, Alon and Goldberg, Yoav},
  title     = {Towards faithfully interpretable {NLP} systems: how should we define and evaluate faithfulness?},
  booktitle = {Proc. ACL},
  year      = {2020}
}

@inproceedings{turpin2023,
  author    = {Turpin, Miles and others},
  title     = {Language models don't always say what they think: unfaithful explanations in chain-of-thought prompting},
  booktitle = {Proc. NeurIPS},
  year      = {2023}
}

@article{lanham2023,
  author  = {Lanham, Tamera and others},
  title   = {Measuring faithfulness in chain-of-thought reasoning},
  journal = {arXiv preprint arXiv:2307.13702},
  year    = {2023}
}

@article{chen2026,
  author  = {Chen, Yu-Wen and Hirschberg, Julia},
  title   = {An audio language model-based voice concept bottleneck framework for interpretable health assessment},
  journal = {arXiv preprint arXiv:2607.16967},
  year    = {2026}
}

@inproceedings{schuller2013,
  author    = {Schuller, Bj{\"o}rn and others},
  title     = {The {INTERSPEECH} 2013 computational paralinguistics challenge: social signals, conflict, emotion, autism},
  booktitle = {Proc. Interspeech},
  pages     = {148--152},
  year      = {2013}
}

@misc{heyjay,
  author       = {Moro-Velazquez, Laureano and others},
  title        = {{HeyJay!}: a corpus of atypical speech for spoken language understanding and automatic speech recognition, {United States}, 2023--2024},
  howpublished = {ICPSR, Version V6},
  year         = {2026},
  doi          = {10.3886/ICPSR39448.V6}
}

@article{nemotronomni,
  author  = {{NVIDIA}},
  title   = {{Nemotron 3 Nano Omni}: efficient and open multimodal intelligence},
  journal = {arXiv preprint arXiv:2604.24954},
  year    = {2026}
}

@article{minicpmo45,
  author  = {{OpenBMB}},
  title   = {{MiniCPM-o 4.5}: towards real-time full-duplex omni-modal interaction},
  journal = {arXiv preprint arXiv:2604.27393},
  year    = {2026}
}

@article{mossaudio,
  author  = {{OpenMOSS Team}},
  title   = {{MOSS-Audio} technical report},
  journal = {arXiv preprint arXiv:2606.01802},
  year    = {2026}
}

@article{gemma4,
  author  = {{Gemma Team}},
  title   = {Gemma 4 technical report},
  journal = {arXiv preprint arXiv:2607.02770},
  year    = {2026}
}

@inproceedings{mmsu,
  author    = {Wang, Dingdong and others},
  title     = {{MMSU}: a massive multi-task spoken language understanding and reasoning benchmark},
  booktitle = {Proc. ICLR},
  year      = {2026}
}

@article{mmarrubrics,
  author  = {Ma, Ziyang and others},
  title   = {The {Interspeech} 2026 audio reasoning challenge: evaluating reasoning process quality for audio reasoning models and agents},
  journal = {arXiv preprint arXiv:2602.14224},
  year    = {2026}
}

@article{audioprocessbench,
  author  = {Zhao, Xiangyu and others},
  title   = {{AudioProcessBench}: benchmark for identifying process errors in audio-grounded reasoning},
  journal = {arXiv preprint arXiv:2606.09925},
  year    = {2026}
}

@article{emosura,
  author  = {Jing, Xin and others},
  title   = {{EmoSURA}: towards accurate evaluation of detailed and long-context emotional speech captions},
  journal = {arXiv preprint arXiv:2603.09820},
  year    = {2026}
}

@article{amiri2026,
  title   = {Augmenting Large Audio Language Models with Low-Level Acoustic Features for Dysarthric Speech Detection},
  author  = {Amiri, Mahdi and Otroshi Shahreza, Hatef and Frossard, Pascal and Kodrasi, Ina},
  journal = {arXiv preprint arXiv:2610.03352},
  year    = {2026}
}

@inproceedings{patel2025,
  title     = {The Sound of Syntax: Finetuning and Comprehensive Evaluation of Language Models for Speech Pathology},
  author    = {Patel, Fagun and others},
  booktitle = {Proc. EMNLP},
  year      = {2025}
}

@inproceedings{chen2026listen,
  title     = {Do Audio {LLMs} Really {LISTEN}, or Just Transcribe? Measuring Lexical vs. Acoustic Emotion Cues Reliance},
  author    = {Chen, Jingyi and others},
  booktitle = {Proc. EACL},
  year      = {2026}
}

\end{document}